\documentclass[runningheads]{llncs}
\usepackage[T1]{fontenc}
\usepackage{graphicx}
\usepackage[table]{xcolor}

\begin{document}
\title{Adapting Classifiers To Changing Class Priors During Deployment}
%
%
\author{Natnael Daba\inst{2}\orcidID{0009-0007-7915-1117} \and
Bruce McIntosh\inst{1} \and
Abhijit Mahalanobis\inst{2}\orcidID{0000-0002-2782-8655}}
%
%
\institute{Department of Computer Science, University of Central Florida, Orlando, FL 32816, USA \and
Department of Electrical and Computer Engineering, University of Arizona, Tucson, AZ 85721, USA}
\maketitle              
\begin{abstract}
Conventional classifiers are trained and evaluated using “balanced” data sets in which all classes are equally present. Classifiers are now trained on large data sets such as ImageNet, and are now able to classify hundreds (if not thousands) of different classes.  On one hand, it is desirable to train such general-purpose classifier on a very large number of classes so that it performs well regardless of the settings in which it is deployed.  On the other hand, it is unlikely that all classes known to the classifier will occur in every deployment scenario, or that they will occur with the same prior probability. In reality, only a relatively small subset of the known classes may be present in a particular setting or environment. For example, a classifier will encounter mostly animals if its deployed in a zoo or for monitoring wildlife, aircraft and service vehicles at an airport, or various types of automobiles and commercial vehicles if is used for monitoring traffic. Furthermore, the exact class priors  are generally unknown and can vary over time. In this paper, we explore different methods for estimating the class priors based on the output of the classifier itself. We then show that incorporating the estimated class priors in the overall decision scheme enables the classifier to increase its run-time accuracy in the context of its deployment scenario.

\keywords{unknown priors \and context \and  classifier \and test time performance accuracy.}
\end{abstract}
\section{Introduction}
Conventional classifiers are trained and evaluated using “balanced” data sets in which all classes are equally present. For instance, well known data sets such as CIFAR-100 and ImageNet (to name a few) have the same number of training and test images for all classes. However, it is unlikely that the prior probability of occurrence will be the same for the classes that are encountered in a given deployment scenario. In fact, many well-known off-the-shelf networks are trained on ImageNet and then deployed in various applications where they will encounter only a relatively small subset of classes. For example, a network trained on ImageNet (with a thousand classes) deployed for traffic surveillance and monitoring is likely to encounter various types of vehicles, pedestrians, cyclists, and some animals such as birds and dogs. However, it is unlikely that most other classes known to the classifier (such as aircraft, large wild animals, marine creatures, and so forth) will be present in such deployment scenarios. Knowledge of the class priors allows the decisions to be weighted in favor of the likely classes, and thereby improves the classifier’s accuracy in a given deployment scenario.

Unfortunately, the exact class priors are generally unknown and can vary over time. In this paper, we explore different methods for estimating the class priors based on the response of the classifier itself, as a means for adapting to unknown or time-varying class mixtures at different sites of operation. We then show that incorporating the estimated class priors in the overall decision scheme (as shown in Figure 1) improves the classifier’s accuracy in the context of the deployment site. It should be noted that no retraining is required, and the method can be applied to any classifier trained on a large number of classes. Results of experiments show that performance can be improved by as much as $10 \%$ for individual classes in many cases.

\begin{figure}
\centering
\includegraphics[width=0.75\textwidth]{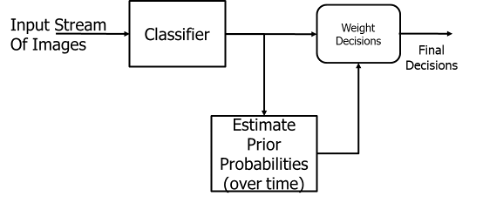}
\caption{The prior probabilities are estimated using the classifier’s decision frequency, and re-weights the decision confidence to improve overall accuracy.} \label{fig1}
\end{figure}

The importance of estimating class priors has been investigated by other researchers. Katti~\cite{harish} states that detecting objects in real world scenes is difficult, and despite huge successes in CNN based detection, humans still generally do better. One proposed reason for this is people understand the context or the environment within which an object is found, and whether a strong correlation exists between the class of the object and context in which it is encountered. Hasan~\cite{hasan} says that in nature, objects and their surroundings tend to coexist in a particular configuration known as a ‘context’ and that visual recognition systems have taken advantage of this to improve results. Aude~\cite{oliva} informs us that objects co-vary with one another, and particular environments create associations that can be exploited by visual system. Sulc~\cite{sulc} points out that a common assumption is that training and test data are independently sampled from an identical distribution; however this is frequently not the case in practice. This discrepancy between training and test time data is posed as a domain adaptation problem, and they propose a method of maximum a posteriori estimation to improve classifier accuracy by learning and using the prior likelihoods of the test set. 

Prior works such as~\cite{saerens,sipka} have explored different test-time prior adaptation methods using a Bayesian framework. For example,~\cite{saerens} proposed an iterative algorithm that is an instance of the expectation-maximization (EM) algorithm for estimating unknown priors at test time. ~\cite{sipka} treats the problem of prior estimation as a maximum likelihood estimation based on a confusion matrix by defining and maximizing a likelihood function of class priors. While we also use a Bayesian framework to formulate the problem, we then show that a naive estimation technique based on the classifier's decision frequency can give good results under certain condition. We also introduce a simpler quadratic programming method for estimating priors that has not been explored in previous works. 

\section{Estimation Of Class Priors Based On Decision Frequency}
 We now discuss several different approaches for estimating the prior probabilities for the different classes using the observed run-time response of the classifier. Assume that $x$  is an image (or data vector) that we seek to classify as one of $K$ possible classes. The true identities of these classes are denoted by $w_{i}$ for $i=1,\ldots,K$. Let $P(w_{i}|x)$ denote the posterior probability for class $i$ conditioned on the observation $x$, and  $P(w_{i})$ be the prior probability of occurrence for class $i$.  By design, the classifier’s output (confidence score) in response to the input data is a proxy for $P(w_{i}|x)$, and will be denoted by $\hat{P} (w_{i}|x)$. The softmax function used at the output of the classifier ensures that $\sum_{i=1}^{K}\hat{P}(w_{i}|x)$, and the classifier decides in favor of class $i$ if $\hat{P}(w_{i}|x) > \hat{P}(w_{j}|x)$ for all $j \neq i$. In doing so, it assumes that the prior likelihood of all classes is the same. Although this is true in a balanced test set, it is unlikely to be valid in a real world setting. 

During training with balanced data, the priors for all classes is the assumed to be the same, say $P = 1/K$, where $K$ is the total number of classes in the dataset. Using Bayes rule, the classifier’s default decision confidence can be interpreted to be  $\hat{P}(w_{i}|x) \cong \frac{P(x|w_{i}) \cdot P}{P(x)}$.  To take the estimated class prior into account when the network is deployed, we multiply the output of the classifier by the scalar ratio $\alpha_{i} = \frac{\hat{P}(w_{i})}{P}$ , where $\hat{P}(w_{i})$ is the estimated class prior. This yields

\begin{equation}
\alpha_{i}\hat{P}(w_{i}|x) \cong \left ( \frac{\hat{P}(w_{i})}{P} \right ) \left ( \frac{p(x|w_{i}) \cdot P}{p(x)} \right ) = \frac{p(x|w_{i}) \cdot \hat{P}(w_{i})}{p(x)}
\label{eq:eqn1}
\end{equation}

which is the corrected estimate for the posterior probability for class $i$ using $\hat{P}(w_{i})$ instead of $P$. Our goal is to then modify the decision strategy to choose class $i$ if $\alpha_{i}\hat{P}(w_{i}|x) > \alpha_{j}\hat{P}(w_{j}|x)$ for all $j \neq i$. However, since $P$ is the same for all classes, this is the same as choosing class $i$ if  $\hat{P}(w_{i}|x)\hat{P}(w_{i}) > \hat{P}(w_{j}|x)\hat{P}(w_{j})$ for all $j \neq i$.

A relatively simple approach for estimating $\hat{P}(w_{i}|x)$ while the classifier is in operation is as follows. Let the decision made by the classifier in response to $x$ be denoted by $C_{j}$, for $j=1, \ldots ,K$. Further, let $P(C_{i}|w_{i})$ represent the probability that the classifier will decide $C_i$  when the true class of $x$ is $w_{i}$. Similarly, the probability that the true class of $x$ is $w_{i}$ when the classifier decides $C_{i}$ is denoted by $P(w_{i}|C_i)$. Using Bayes rule, we obtain $P(w_{i}|C_{i}) = \frac{P(C_{i}|w_{i})P(w_{i})}{P(C_{i})}$
where $P(w_{i})$ is the yet unknown prior probability of class $i$. Since  $P(C_{i}|w_{i})$ is the same as \textbf{recall}, and  $P(w_{i}|C_{i})$ is the same as \textbf{precision}, we can re-write this  as 


\begin{equation}
    P(w_{i}) = \frac{P(w_{i}|C_{i})P(C_{i})}{P(C_{i}|w_{i})} = \frac{Precision_{i}\times P(C_{i})}{Recall_{i}}
    \label{eq:eqn3}
\end{equation}

It should be noted that $P(C_{i})$ is the overall probability that the classifier decides class $i$ regardless of what class the input $x$ belongs to. To estimate this term, assume that the classifier is observed to process a total of $N$ test images during operation, and $N_{i}$ are the number of them classified as belonging to class $i$. A rough estimate for $P(C_i )$ is then given by  $\hat{P}(C_{i})=N_{i}/N$ so that the prior probability for class $i$ is estimated as

\begin{equation}
    \hat{P}(w_{i}) = \frac{Precision_{i} \times N_{i}}{Recall_{i} \times N} 
\end{equation}

Of course, the classifier’s precision and recall for the different classes can be readily obtained from the “off-line” evaluations on balanced test sets. If precision and recall are approximately equal, we get $\hat{P}(w_{i}) = \frac{N_{i}}{N}$. 
We will use this as a “naïve estimate” to modify the decision strategy for the classifier such that it decides class $i$ if  $\hat{P}(w_{i}|x)\hat{P}(w_{i}) > \hat{P}(w_{j}|x)\hat{P}(w_{j})$ for all $j \neq i$. This re-weights the classifier’s default decision confidences (when deployed) using the naïve estimate of the class priors, and thereby adapts its performance to the chances of encountering the various classes in the environment in which it is operating.

\textbf{Quadratic Programming approach:} The previous naïve estimate does not take joint error probabilities into account, and may not provide optimum results if the incorrect decisions are highly correlated, or when the precision and recall differ significantly. To address this problem, consider a deployed classifier which produces a sequence of decisions in response to an input stream of images. Assume that these images belong to a subset of the classes known to the classifier, but the prior probability of occurrence of each class is not known. We also assume that the performance of the classifier on a balanced test set is available in the form of a confusion matrix, where each row has been normalized to add to $1.0$. We can then treat the entry in the $j$-th row and $i$-th column as an estimate of $P(C_{i} | w_{j})$, i.e. the probability that the classifier’s decision is $C_{i}$ when the true class is $w_{j}$. By definition, we know that $P(C_{i}) = \sum_{j=1}^{K} P(C_{i} | w_{j})P(w_{j})$, where $K$ is the total number of classes known to the classifier.  However, $P(C_{i})$ is nothing but the frequency of the decisions made by the classifier, and is essentially the normalized histogram of the classifier’s decisions over time, which can be observed and measured in the deployed environment. What is not known are the prior probabilities $P(w_{j})$.


To solve for $P(w_{j})$, consider the $j$-th row of the normalized confusion matrix represented as a column vector $\textbf{h}_{j} = \left [\: P(C_{1} |w_{j}) \: \: P(C_{2}|w_{j}) \: \: \ldots \: \: P(C_{k} |w_{j}) \right ]^{T}$
These can be arranged as the columns of the matrix $\textbf{H} = \left [ \textbf{h}_{1} \: \: \textbf{h}_{2} \: \: \ldots \: \: \textbf{h}_{k} \right]$. We also express the histogram of the observed decisions $P (C_{i})$, for $i=1, ...,  k$ as the vector $ \textbf{c} = \left [\: P(C_{1}) \: \: P(C_{2}) \: \: \ldots \: \: P(C_{k}) \right ]^{T}$,
and the vector of unknown class priors as $\textbf{v} = \left [\: P(w_{1}) \: \: P(w_{2}) \: \: \ldots \: \: P(w_{K}) \right ]^{T}$.
                                                                                                                            
Using these definitions, the equation $P(C_{i}) = \sum_{j=1}^{K} P(C_{i} | w_{j})p(w_{j})$ can then be written simply as $\textbf{c} = \textbf{Hv}$.  Note that $\textbf{H}$ is known apriori from the training process, and is a normalized version of the confusion matrix. Similarly, $\textbf{c}$ is the histogram of the observed decisions when the network is deployed.  A simple solution for $\textbf{v}$ is of course obtained by inverting the matrix $\textbf{H}$ (i.e $\textbf{v}=\textbf{H}^{-1}\textbf{c}$) but this does not guarantee positive values between 0 and 1 for the elements of $\textbf{v}$.  Therefore, we use quadratic programming to find $\textbf{v}$  by solving the problem
\begin{equation}
    minimize \: \: (\textbf{Hv}-\textbf{c})^{2} \: \: subject \: \: to \sum_{k=1}^{K} v(k) = 1.0 \: \: and \: \: 0 \leq v(k) \leq 1.0,
\end{equation}
where $v(k)$ (or equivalently $P(w_{k})$) is an element of \textbf{v}. The resulting values are the estimates of the prior probabilities of the classes that are present in the input stream.

\section{Method and Results}
We now apply this idea to the CIFAR-100 [5] and Tiny ImageNet [6] datasets using the Resnet18 [7] and Densenet201 [8] classifiers. Both classifiers are pretrained on ImageNet, but finetuned on CIFAR-100 and Tiny ImageNet (TIN). The first experiment is conducted on the CIFAR-100 dataset. To simulate different deployment scenarios, we define twelve “super-classes” or “context scenarios” which contain from 5 to 25 classes each. These included aquatic animals, food containers, flora, electrical items, fruits and vegetables, furniture, insects, man-made things, animals, people, outdoor places, and vehicles. The premise is that although the classifier knows a hundred classes, it will only encounter a relatively small number of them when operating in any one of these settings. For example, when the classifier is deployed to recognize aquatic animals, it will not be presented with fruits and vegetables and so forth. However, the classifier does not know which contextual scenario it is operating in. The goal is to estimate the class priors using the method described in the previous section, and employ it to increase the classifier’s accuracy in the deployed setting.

The CIFAR-100 data set provides 100 test images per class. Of these, 50 test images per class were used for estimating the matrix \textbf{H} (which is based on the normalized confusion matrix).  We also create a transfer set consisting of 20 samples of each class drawn from the test set. The main purpose of this set is to estimate the histogram of the decisions for each context scenario, which then provides the vector \textbf{c}.  We then solve for $\textbf{v}$ using the techniques proposed in Section 2, and use it to weight the classifier’s outputs. Finally, the remaining 30 images for each class are used to evaluate the decision strategy and assess the impact of the proposed approach on the overall accuracy of the classifier.  We are careful to make sure that the three splits of the test data are exclusive of one another. As noted previously, the simple matrix inverse results in negative values for some of the class priors, so these are mapped to 0. The quadratic programming approach constrains the resulting priors on the closed interval [0,1]. The CVXPY convex optimization library was used for this task. 
\begin{table}
\caption{CIFAR-100/ResNet-18: Comparison of classifier accuracy using different techniques for estimating class priors. Using the estimated priors always achieves higher accuracy than the baseline }\label{tab1}
\centering 
\resizebox{\textwidth}{!}{%
\begin{tabular}{|c|c|c|c|c|c|c|c|c|c|c|c|c|}
\hline
Class ids & 0 & 1 & 2 & 3 & 4 & 5 & 6 & 7 & 8 & 9 & 10 & 11 \\
\hline
\textbf{Baseline} & 0.643 & 0.683 & 0.725 & 0.736 & 0.799 & 0.773 & 0.729 & 0.822 & 0.687 & 0.573 & 0.763 & 0.806 \\
\hline 
\textbf{Naive} & $\textbf{0.733}$ & 0.785 & 0.778 & 0.855 & 0.859 & 0.827 & 0.788 & 0.9 & 0.725 & \textbf{0.585} & 0.856 & 0.844 \\
\hline
\textbf{Matrix inverse} & 0.722 & 0.803 & \textbf{0.799} & 0.854 & \textbf{0.867} & 0.833 & \textbf{0.81} & \textbf{0.909} & 0.733 & 0.524 & \textbf{0.859} & 0.854 \\
\hline
\textbf{Quadratic programming} & 0.731 & \textbf{0.805} & \textbf{0.799} & \textbf{0.863} & \textbf{0.867} & \textbf{0.838} & 0.809 & 0.908 & \textbf{0.735} & 0.57 & \textbf{0.859} & \textbf{0.856} \\
\hline
\rowcolor{lightgray} \textbf{Ground truth} & 0.751 & 0.818 & 0.798 & 0.875 & 0.873 & 0.841 & 0.81 & 0.907 & 0.759 & 0.614 & 0.867 & 0.854 \\
\hline
\end{tabular}%
}
\end{table}
\vspace{-30pt}

\begin{table}
\caption{CIFAR-100/DenseNet-201}\label{tab1}
\resizebox{\textwidth}{!}{%
\begin{tabular}{|c|c|c|c|c|c|c|c|c|c|c|c|c|}
\hline
Class ids & 0 & 1 & 2 & 3 & 4 & 5 & 6 & 7 & 8 & 9 & 10 & 11 \\
\hline
\textbf{Baseline} & 0.751 & 0.807 & 0.859 & 0.853 & 0.923 & 0.863 & 0.842 & 0.897 & 0.847 & 0.633 & 0.843 & 0.92 \\
\hline 
\textbf{Naive} & \textbf{0.744} & \textbf{0.793} & \textbf{0.861} & 0.833 & 0.937 & 0.9 & 0.871 & 0.9 & 0.84 & \textbf{0.657} & 0.873 & \textbf{0.927} \\
\hline
\textbf{Matrix inverse} & 0.732 & 0.783 & 0.824 & \textbf{0.913} & \textbf{0.94} & \textbf{0.903} & 0.867 & \textbf{0.923} & 0.84 & 0.58 & \textbf{0.88} & \textbf{0.927} \\
\hline
\textbf{Quadratic programming} & 0.732 & 0.78 & 0.824 & \textbf{0.913} & 0.937 & \textbf{0.903} & \textbf{0.881} & \textbf{0.923} & \textbf{0.848} & 0.577 & \textbf{0.88} & \textbf{0.927} \\
\hline
\rowcolor{lightgray} \textbf{Ground truth} & 0.818 & 0.86 & 0.887 & 0.933 & 0.95 & 0.923 & 0.923 & 0.937 & 0.875 & 0.69 & 0.913 & 0.935 \\
\hline
\end{tabular}%
}
\end{table}

The results of different comparisons are shown in Table 1 for ResNet18, and in Table 2 for DenseNet201. Since the number of samples to work with for these experiments is relatively small, we make use of 10-fold cross-validation where each fold is a different mixture of samples between the context transfer and context test sets.  The columns represent a particular “context” or scenario, where the test images are restricted to one of the 12 super-classes. There are five scores for every scenario, each represented by one of the rows of the table.  In any given column, the first row is the baseline performance obtained using the default score of the classifier without using any estimated priors. The bottom row is the “best possible” or ideal score that could be obtained if the exact true priors were known. The three rows in-between represent the naïve estimate, the matrix inverse solution, and the priors obtained using quadratic programming, respectively.  The largest among these is indicated by the bold font. It is clear from these results that in every scenario, using the estimated priors yields higher accuracy than the baseline.  We see that for ResNet-18, the proposed techniques can improve accuracy by as much as $12 \%$ compared to the baseline (scenario 1), in some cases achieving the theoretical maximum improvement (i.e. comparable to the last row). The quadratic programming approach achieves the best result in eight of the twelve scenarios. For DenseNet201, the distinction between the different technique is not as much, which indicates that precision and recall are comparable for this network. The quadratic programming approach gives the best result in seven of the scenarios, and the estimated priors improve the accuracy compared to the baseline in all cases except for scenarios 0 and 1.

The second experiment is conducted on the Tiny ImageNet (TIN) dataset.  This dataset contains 64x64 RGB images of 200 classes with 500 images per class for the training set, 50 images per class for the validation set, and 50 images per class for the test set. For this experiment, we selected 36 classes (out of the entire 200 classes) of the dataset, and created twelve contexts/superclasses each with 3 member classes. The groupings are as follows: amphibian, arthropod, bird, car, cat, clothing, cooking utensil, dog, edible fruit, furniture, geological formation, and timepiece. The transfer set is made up of 20 samples of each member class and thus contains a total of 12x3x20 = 720 images. The test, which is separate from the transfer set, is made up of 30 samples of each member class and thus contains a total of 12x3x30 = 1080 images. 

Once again, we used ResNet-18 and DenseNet-201 that are pretrained on 224x224 sized images from the ImageNet-1k dataset. These models are then fine-tuned on the 64x64 sized images from the TIN dataset. The results are shown in Table 3 and 4, which are all results are obtained using 10-fold cross validation.  
 In all cases, using the estimated class priors improves the accuracy of the classifier compared to the baseline, sometimes by as much  15\% for ResNet-18 (e.g. scenario 6), and by more than 10\% for DenseNet-201 (e.g. Scenario 7). In some cases, this improvement approaches the theoretical best possible performance (last row of the table). While all methods for estimating the priors result in improvement in accuracy, the quadratic optimization yields the most improvement compared to the naïve method and the matrix inverse solution for both networks. As a result of incorporating the quadratic programming method into our algorithm, we identified a computational overhead characterized by a time complexity of $O(n^{3.5})$ (incurred because of the quadratic programming solver method) and a space complexity of $O(n^{2})$ (incurred when storing the confusion matrix) where $n$ denotes the number of classes.


\begin{table}
\caption{Tiny-ImageNet/ResNet-18}\label{tab1}
\resizebox{\textwidth}{!}{%
\begin{tabular}{|c|c|c|c|c|c|c|c|c|c|c|c|c|}
\hline
Class ids & 0 & 1 & 2 & 3 & 4 & 5 & 6 & 7 & 8 & 9 & 10 & 11 \\
\hline
\textbf{Baseline} & 0.718 & 0.829 & 0.871 & 0.656 & 0.656 & 0.714 & 0.652 & 0.721 & 0.777 & 0.741 & 0.698 & 0.817 \\
\hline 
\textbf{Naive} & 0.819 & 0.918 & 0.931 & \textbf{0.793} & 0.762 & 0.903 & 0.804 & 0.81 & 0.882 & 0.916 & 0.809 & \textbf{0.959} \\
\hline
\textbf{Matrix inverse} & 0.831 & 0.923 & \textbf{0.932} & 0.772 & 0.78 & 0.919 & 0.819 & 0.819 & 0.913 & \textbf{0.93} & 0.828 & 0.951 \\
\hline
\textbf{Quadratic programming} & \textbf{0.833} & \textbf{0.93} & \textbf{0.932} & 0.78 & \textbf{0.781} & \textbf{0.922} & \textbf{0.82} & \textbf{0.82} & \textbf{0.914} & \textbf{0.93} & \textbf{0.832} & 0.958 \\
\hline
\rowcolor{lightgray} \textbf{Ground truth} & 0.844 & 0.948 & 0.93 & 0.803 & 0.761 & 0.946 & 0.818 & 0.813 & 0.936 & 0.931 & 0.832 & 0.954 \\
\hline
\end{tabular}%
}
\end{table}
\vspace{-30pt}
\begin{table}
\caption{Tiny-ImageNet/DenseNet-201}\label{tab1}
\resizebox{\textwidth}{!}{%
\begin{tabular}{|c|c|c|c|c|c|c|c|c|c|c|c|c|}
\hline
Class ids & 0 & 1 & 2 & 3 & 4 & 5 & 6 & 7 & 8 & 9 & 10 & 11 \\
\hline
\textbf{Baseline} & 0.78 & 0.834 & 0.898 & 0.687 & 0.679 & 0.82 & 0.777 & 0.738 & 0.847 & 0.816 & 0.7 & 0.877 \\
\hline 
\textbf{Naive} & \textbf{0.867} & 0.969 & 0.92 & 0.796 & 0.768 & 0.949 & \textbf{0.848} & 0.844 & 0.942 & 0.953 & 0.837 & 0.977 \\
\hline
\textbf{Matrix inverse} & \textbf{0.867} & 0.973 & \textbf{0.924} & 0.824 & 0.787 & \textbf{0.954} & 0.84 & 0.859 & 0.96 & 0.958 & 0.841 & 0.973 \\
\hline
\textbf{Quadratic programming} & \textbf{0.867} & \textbf{0.977} & \textbf{0.924} & \textbf{0.826} & \textbf{0.789} & \textbf{0.954} & 0.84 & \textbf{0.863} & \textbf{0.96} & \textbf{0.959} & \textbf{0.842} & \textbf{0.98} \\
\hline
\rowcolor{lightgray} \textbf{Ground truth} & 0.873 & 0.98 & 0.926 & 0.838 & 0.781 & 0.954 & 0.85 & 0.869 & \textbf{0.96} & 0.953 & 0.868 & 0.989 \\
\hline
\end{tabular}%
}
\end{table}
\section{Conclusion}
The proposed techniques enables the classifier to adapt to the unknown class prior probabilities in the deployment environment, which are likely to vary by location and over time. Specifically, we showed that these can be estimated in a Bayesian framework using the histogram of a classifier’s decisions observed over a period of time. Experiments using the CIFAR-100 and Tiny ImageNet data set using ResNet18 and DenseNet201 show that in some instances this can lead to as much as $15\%$ improvement in performance compared to the classifier's baseline accuracy. On the CIFAR100 dataset with ResNet-18, we observe that in almost all contexts (except for 0 and 9), the matrix inverse and quadratic programming methods yield better performance in terms of classification accuracy. Regarding the performance of DenseNet-201 on the CIFAR-100, there are 5 contexts (0,1,2,9 and 11) for which the naive approach yields a slightly better result than the matrix inverse and quadratic programming approaches.  For these cases, we have observed that the precision and recall of the classifier are approximately the same, which makes the naïve estimate more accurate. Overall, across all experiments, the quadratic programming approach offers the greatest improvement in classification accuracy (in 36 of the 48 scenarios in all four tables).

%
%
%
%





\end{document}